# AN OPTIMAL CONSENSUS TRACKING CONTROL ALGORITHM FOR AUTONOMOUS UNDERWATER VEHICLES WITH DISTURBANCES

Jian Yuan[1] Wen-Xia Zhang[2] and Zhou-Hai Zhou[1]

[1]Institute of Oceanographic Instrumentation of Shandong Academy of Science, Qingdao, China

jyuanjian801209@163.com

[2] Mechanical & Electrical Engineering Department, Qingdao College, Qingdao,China

dasinusheng@126.com

## ABSTRACT

*The optimal disturbance rejection control problem is considered for consensus tracking systems affected by external persistent disturbances and noise. Optimal estimated values of system states are obtained by recursive filtering for the multiple autonomous underwater vehicles modeled to multi-agent systems with Kalman filter. Then the feedforward-feedback optimal control law is deduced by solving the Riccati equations and matrix equations. The existence and uniqueness condition of feedforward-feedback optimal control law is proposed and the optimal control law algorithm is carried out. Lastly, simulations show the result is effectiveness with respect to external persistent disturbances and noise.*

## KEYWORDS

*Autonomous Underwater Vehicles; Consensus Tracking; Optimal Disturbance Rejection*

## 1. INTRODUCTION

The AUVs formation control is a typical problem of multi-robot coordination and cooperation. The coordinated control of multiple AUVs can significantly improve many applications including ocean sampling, imaging, and surveillance abilities. The large-scale multiple AUVs system is modelled to multi-agent system to study the consensus problem on their spatial location. Wang and Xiao [1] proposed a finite-time formation control framework for large-scale multi -agent system. They divide the formation information into two types: global information and local information, in which the global information can decide the formation shape and only the leader can obtain such the global information; followers can only get local information. This framework can reduce the amount of communication between the agents. And then they design a nonlinear consensus protocol, and apply it to the time-invariant, time-varying and trajectory-tracking control. Wang and Hong [2] propose some types of consensus control algorithms for the first-order dynamic system with variable coupling topology. They design a finite-time consensus protocol and give its non-smooth controller using time-invariant Lyapunov function and graph theory tool. Further, they propose a non-smooth time-invariant consensus algorithm for a second-order dynamic systems. They demonstrate the existence of the finite-time control law using Lyapunov functions and graph theory in [3]. Wang and Chen [4] study the consensus problem of continuous-time multi-agent with communication time delay. They design a class of continuous but non-smooth finite-time controller which ensures that the multi-agent systems with time delay reach a consistent state in finite time. The underwater environment is extremely complex, of strong noise and disturbance, resulting in multi-agent status subjecting to these external disturbances. Reducing or overcoming the impact of these disturbances and noise on multi-agent formation has important theoretical and practical significance. In this paper, the feedforward-feedback optimal tracking control problem for the multi-agent system with external disturbance and noise under a given performance index based on the Kalman filter is studied. Optimal estimated values of system states are obtained by recursive filtering for the multiple autonomous underwater vehicles modelled to multi-agent systems with Kalman filter. Then the feedforward-feedback optimal control law is deduced by solving the Riccati equations and matrix equations. The existence and uniqueness condition of feedforward-feedback optimal control law is proposed and the optimal control law algorithm is carried out. Simulations show the result is effectiveness with respect to external persistent disturbances and noise.

## 2. PROBLEM DESCRIPTION

The consensus algorithm with external disturbance is described as

$$\dot{x}_i(t) = \sum_{v_j \in N_i^t} a_{ij}(x_j(t) - x_i(t)) + a_{vi}v_i(t) + a_{i0}(x_0(t) - x_i(t)) + m_i(t) \qquad (1)$$

where $v_i(t)$ denotes external disturbance and satisfies $\dot{v}_i(t) = f(v_i(t))$, $x_0(t)$ is the desired trajectory and satisfies $\dot{x}_0(t) = f(x_0(t))$, $m_i(t)$ denotes Gaussian white noise where $a_{i0} > 0$, $a_{vi} > 0$. without loss of generality, the individual AUV dynamics of multi-AUV systems with disturbances described as follows:

$$\dot{x}_i(t) = A_{ii}x_i(t) + \sum_{j \neq i} A_{ij}x_j(t) + B_{1i}u_i(t) + B_{2i}w_i(t) + m_i(t)$$

$$y_i(t) = C_ix_i(t) + n_i(t) \qquad\qquad , i \in N \qquad (2)$$

$$x_i(0) = x_{i0}$$

where $x_i(t)$ denotes the states of AUV $i$ and $x_i(t) \in R^p$. $x_j(t)$ denotes the states of AUV $j$ surrounding AUV $i$. $y_i(t) \in R^s$ denotes the system output. $u_i(t) \in R^q$ denotes the input of AUV $i$.

$w_i(t) \in R^r$ denotes the disturbances, $m_i(t) \in R^p$ denotes the process noise, $n_i(t) \in R^p$ denotes the measurement noise. $A_{ii} \in R^{p \times p}$, $A_{ij} \in R^{p \times p}$, $B_{1i} \in R^{p \times q}$, $B_{2i} \in R^{p \times r}$ and $C_i \in R^{s \times p}$ are constant matrice of appropriate dimensions. $A_{ij}$ denotes the corresponding matrix between AUV $i$ and AUV $j$.

The output trajectory-tracking $\tilde{y}_i$ of virtual leader which the system output $y_i$ tracks is described as

$$\dot{\boldsymbol{z}}_i(t) = \boldsymbol{F}_i \boldsymbol{z}_i(t) + \boldsymbol{m}_{zi}(t)$$
$$\tilde{\boldsymbol{y}}_i(t) = \boldsymbol{H}_i \boldsymbol{z}_i(t) + \boldsymbol{n}_{zi}(t)$$

$$(3)$$

where $\boldsymbol{z}_i \in R^l$, $\tilde{\boldsymbol{y}}_i \in R^s$, $\boldsymbol{F}_i \in R^{l \times l}$ and $\boldsymbol{H}_i \in R^{s \times l}$ are constant matrice of appropriate dimensions. And $(\boldsymbol{F}_i, \boldsymbol{H}_i)$ is observable. $\boldsymbol{m}_{zi}(t) \in {}^{l}$ denotes the process noise of the external system, $\boldsymbol{n}_{zi}(t) \in R^s$ is the measurement noise of the external system. So the equation (2) is rewritten as

$$\dot{\boldsymbol{x}}(t) = \boldsymbol{A}\boldsymbol{x}(t) + \boldsymbol{B}_1\boldsymbol{u}(t) + \boldsymbol{B}_2\boldsymbol{w}(t) + \boldsymbol{m}(t)$$
$$\boldsymbol{y}(t) = \boldsymbol{C}\boldsymbol{x}(t) + \boldsymbol{n}(t)$$
$$\boldsymbol{x}(0) = \boldsymbol{x}_0$$

$$(4)$$

where

$$\boldsymbol{x}(t) = \begin{bmatrix} \boldsymbol{x}_1^{\mathrm{T}} & \boldsymbol{x}_2^{\mathrm{T}} & \cdots & \boldsymbol{x}_N^{\mathrm{T}} \end{bmatrix}^{\mathrm{T}} \in R^{Np}, \ \boldsymbol{m}(t) = \begin{bmatrix} \boldsymbol{m}_1^{\mathrm{T}} & \boldsymbol{m}_2^{\mathrm{T}} & \cdots & \boldsymbol{m}_N^{\mathrm{T}} \end{bmatrix}^{\mathrm{T}} \in R^{Np}$$

$$\boldsymbol{u}(t) = \begin{bmatrix} \boldsymbol{u}_1^{\mathrm{T}} & \boldsymbol{u}_2^{\mathrm{T}} & \cdots & \boldsymbol{u}_N^{\mathrm{T}} \end{bmatrix}^{\mathrm{T}} \in R^{Nq}, \ \boldsymbol{w}(t) = \begin{bmatrix} \boldsymbol{w}_1^{\mathrm{T}} & \boldsymbol{w}_2^{\mathrm{T}} & \cdots & \boldsymbol{w}_N^{\mathrm{T}} \end{bmatrix}^{\mathrm{T}} \in R^{Nr}$$

$$\boldsymbol{y}(t) = \begin{bmatrix} \boldsymbol{y}_1^{\mathrm{T}} & \boldsymbol{y}_2^{\mathrm{T}} & \cdots & \boldsymbol{y}_N^{\mathrm{T}} \end{bmatrix}^{\mathrm{T}} \in R^{Ns}, \ \boldsymbol{n}(t) = \begin{bmatrix} \boldsymbol{n}_1^{\mathrm{T}} & \boldsymbol{n}_2^{\mathrm{T}} & \cdots & \boldsymbol{n}_N^{\mathrm{T}} \end{bmatrix}^{\mathrm{T}} \in R^{Ns}$$

$$\boldsymbol{A} = \begin{bmatrix} \boldsymbol{A}_{11} & \cdots & \boldsymbol{A}_{1N} \\ \vdots & \ddots & \vdots \\ \boldsymbol{A}_{N1} & \cdots & \boldsymbol{A}_{NN} \end{bmatrix} \in R^{Np \times Np}, \ \boldsymbol{B}_1 = \begin{bmatrix} \boldsymbol{B}_{11} & & \\ & \ddots & \\ & & \boldsymbol{B}_{1N} \end{bmatrix} \in R^{Np \times Nq}$$

$$\boldsymbol{B}_2 = \begin{bmatrix} \boldsymbol{B}_{21} & & \\ & \ddots & \\ & & \boldsymbol{B}_{2N} \end{bmatrix} \in R^{Np \times Nr}, \ \boldsymbol{C} = \begin{bmatrix} \boldsymbol{C}_1 & & \\ & \ddots & \\ & & \boldsymbol{C}_N \end{bmatrix} \in R^{Ns \times Np}$$

We define the disturbance $\boldsymbol{w}(t)$ as

$$\dot{\boldsymbol{w}}(t) = K\boldsymbol{w}(t) + \boldsymbol{m}_w(t)$$

$$(5)$$

The output trajectory-tracking $\tilde{\boldsymbol{y}}$ of virtual leader which the system output $\boldsymbol{y}$ tracks is described as

$$\dot{\boldsymbol{z}}(t) = \boldsymbol{F}\boldsymbol{z}(t) + \boldsymbol{m}_z(t)$$
$$\tilde{\boldsymbol{y}}(t) = \boldsymbol{H}\boldsymbol{z}(t) + \boldsymbol{n}_z(t)$$

$$(6)$$

where $\boldsymbol{z} \in R^{Nl}$, $\boldsymbol{m}_z(t) \in R^{Nl}$, $\tilde{\boldsymbol{y}} \in R^{Ns}$, $\boldsymbol{n}_z(t) \in R^{Ns}$, $\boldsymbol{F} \in R^{Nl \times Nl}$ and $\boldsymbol{H} \in R^{Ns \times Nl}$ are constant matrices of appropriate dimensions. And $(\boldsymbol{F}, \boldsymbol{H})$ is observable.

The filtered system by optimal state filtering is described as

$$\dot{\hat{x}}(t) = A\hat{x}(t) + B_1\hat{u}(t) + B_2\hat{w}(t)$$
$$\hat{y}(t) = C\hat{x}(t)$$
$$\hat{x}(0) = x_0 \tag{7}$$

where $\hat{x}(t) = \left[\hat{x}_1^{\mathrm{T}} \quad x_2^{\mathrm{T}} \quad \cdots \quad \hat{x}_N^{\mathrm{T}}\right]^{\mathrm{T}} \in R^{Np}, \hat{u}(t) = \left[\hat{u}_1^{\mathrm{T}} \quad u_2^{\mathrm{T}} \quad \cdots \quad \hat{u}_N^{\mathrm{T}}\right]^{\mathrm{T}} \in R^{Nq}$

$\hat{w}(t) = \left[\hat{w}_1^{\mathrm{T}} \quad w_2^{\mathrm{T}} \quad \cdots \quad \hat{w}_N^{\mathrm{T}}\right]^{\mathrm{T}} \in R^{Nr}, \hat{y}(t) = \left[\hat{y}_1^{\mathrm{T}} \quad y_2^{\mathrm{T}} \quad \cdots \quad \hat{y}_N^{\mathrm{T}}\right]^{\mathrm{T}} \in R^{Ns}$

The external disturbance system is described as

$$\dot{\hat{w}}(t) = K\hat{w}(t) \tag{8}$$

The external trajectory-tracking system is described as

$$\dot{\hat{z}}(t) = F\hat{z}(t)$$
$$\dot{\bar{y}}(t) = H\hat{z}(t) \tag{9}$$

where $\hat{z} \in R^{Nl}, \bar{y} \in R^{Ns}$. Choose Infinite horizon quadratic performance index as

$$J = E\left[\int_0^\infty \left(e^{\mathrm{T}}(t)Qe(t) + u^{\mathrm{T}}(t)Ru(t)\right)\mathrm{d}t\right]$$
$$= \int_0^\infty \left(\hat{e}^{\mathrm{T}}(t)Q\hat{e}(t) + \hat{u}^{\mathrm{T}}(t)R\hat{u}(t)\right)\mathrm{d}t \tag{10}$$

where $Q$ and $R$ are positive definite matrix of appropriate dimensions. $e(t)$ and $\hat{e}(t)$ are output error and estimated value of the output error, described as

$$e(t) = \bar{y}(t) - y(t)$$
$$\hat{e}(t) = E[e(t)] = \dot{\bar{y}}(t) - \hat{y}(t) \tag{11}$$

The optimal formation control is to solve the optimal tracking control law $\hat{u}^*(t)$ to make $J$ obtain the minimal value.

## 3. DESIGN ON OPTIMAL FORMATION CONTROL

In this section we focus on the designing on the optimal formation control law of AUVs with disturbance and noise effects. First, we give the following theorem:

Theorem 1 Considered the optimal tracking control problem of disturbed multiple AUVs system with equation (4) and (6) under the performance indicators (10), the optimal formation control law only exists and is represented by the following formula:

$$\hat{u}^*(t) = -R^{-1}B_1^T\left(P\hat{x}(t) + P_1\hat{z}(t) + P_2\hat{w}(t)\right) \tag{12}$$

where $\boldsymbol{P}$ is the unique semi-positive definite solution of the following matrix algebraic equation:

$$A^{\mathrm{T}}P + PA - PSP + C^{\mathrm{T}}QC = 0 \qquad (13)$$

where $\boldsymbol{P_1}$ is the unique solution of the following matrix algebraic equation:

$$P_1 F - PSP_1 - C^{\mathrm{T}}QH + A^T P_1 = 0 \qquad (14)$$

where $\boldsymbol{P_2}$ is the unique solution of the following matrix algebraic equation:

$$A^{\mathrm{T}}P_2 + P_2 K - PSP_2 + PB_2 = 0 \qquad (15)$$

where $\boldsymbol{S} = \boldsymbol{B_1} \boldsymbol{R}^{-1} \boldsymbol{B_1}^{\mathrm{T}}$.

Proof: A necessary condition of the optimal formation control under performance indicators (10) for system (4) and (6)for leads to solving the following two point boundary value:

$$
\begin{aligned}
& -\dot{l}(t) = C^{\mathrm{T}}QC\hat{x}(t) - C^{\mathrm{T}}QH\hat{z}(t) + A^{\mathrm{T}}\boldsymbol{\lambda}(t) \\
& \dot{\hat{x}}(t) = A\hat{x}(t) - S\boldsymbol{\lambda}(t) + B_2\hat{w}(t), \quad t_0 < t < \yen \\
& \boldsymbol{\lambda}(\yen) = \boldsymbol{0} \\
& \hat{x}(0) = \boldsymbol{x_0}
\end{aligned}
\qquad (16)
$$

Its optimal formation control law is described as

$$\hat{u}(t) = -\boldsymbol{R}^{-1}\boldsymbol{B_1}^{\mathrm{T}}\boldsymbol{\lambda}(t) \qquad (17)$$

For solving the two point boundary value(16), we define

$$\boldsymbol{\lambda}(t) = \boldsymbol{P}\hat{x}(t) + \boldsymbol{P_1}\hat{z}(t) + \boldsymbol{P_2}\hat{w}(t) \qquad (18)$$

Take the derivative of both sides of(17), and take the second equation of (16) and (5) into account, we obtain

$$\dot{\boldsymbol{\lambda}}(t) = \dot{P}\hat{x}(t) + P\dot{\hat{x}}(t) + \dot{P_1}\hat{z}(t) + P_1\dot{\hat{z}}(t) + \dot{P_2}\hat{w}(t) + P_2\dot{\hat{w}}(t) \qquad (19)$$

Compare (19) with the first equation of(16), we obtain:

$$
\begin{aligned}
& \left(A^{\mathrm{T}}P + PA - PSP + C^{\mathrm{T}}QC\right)\hat{x}(t) + \left(P_1 F - PSP_1 - C^{\mathrm{T}}QH + A^T P\right)\hat{z}(t) \\
& + \left(A^{\mathrm{T}}P_2 + P_2 K - PSP_2 + PB_2\right)\hat{w}(t) = \boldsymbol{0}
\end{aligned}
\qquad (20)
$$

For all $\hat{x}(t)$, $\hat{z}(t)$ and $\hat{w}(t)$, (19) establishes all the time. So we obtain the *Ricatti* matrix algebraic equation (13), (14) and (15)For equation (13) is the matrix algebraic equation about $\boldsymbol{P}$ with unique semi-positive definite solution, we take it into matrix algebraic equation(14)and(15) to solve the unique $\boldsymbol{P}_1$ and $\boldsymbol{P}_2$ .

When the unique $\boldsymbol{P}$ , $\boldsymbol{P}_1$ and $\boldsymbol{P}_2$ are obtained, $\boldsymbol{\lambda}(t)$ is solved. Furthermore, by (18) we determine the optimal formation control law(12).

For the optimal tracking control problem described by (4) and(6), the designing procedure of the above algorithm is as follows:

**Algorithm:** (the optimal consensus tracking control algorithm)

  Solving the expected output $\hat{\bar{\boldsymbol{y}}}(t)$ through(9);

  Solving $\boldsymbol{P}(t)$ , $\boldsymbol{P}_1(t)$ and $\boldsymbol{P}_2(t)$ through equations(13), (14) and (15), respectively;

  Calculating $\hat{\boldsymbol{x}}(t)$ through(16);

  Calculating $\hat{\boldsymbol{u}}(t)$ through(12);

  Calculating $\hat{\boldsymbol{e}}(t)$ through(11);

  Calculating $J$ through(10).

According to the above system description and optimal consensus tracking control algorithm, we obtain the control block of the optimal consensus tracking control illustrated in Fig.1.

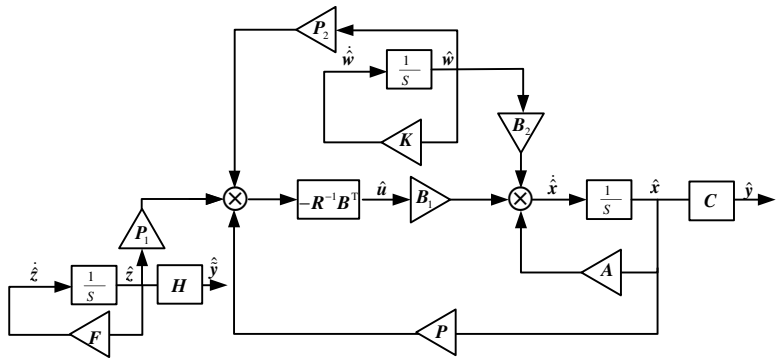

Figure 1. System block of optimal consensus tracking control system with disturbance and noise

## 4. SIMULATIONS

Considering the multiple AUVs system with noise and disturbance described by equations(4), (5) and (6) as follows:

$$\begin{cases} \dot{x}(t) = \begin{bmatrix} -1 & 0 \\ 0 & -1 \end{bmatrix} x(t) + \begin{bmatrix} 3 & 2 \\ 4 & 5 \end{bmatrix} u(t) + \begin{bmatrix} -0.1 & 0 \\ 0 & 1 \end{bmatrix} w(t) + m(t), \ x(0) = \begin{bmatrix} 0 & 0 \end{bmatrix}^{\mathrm{T}} \\ y(t) = \begin{bmatrix} 1 & 2 \\ 3 & 4 \end{bmatrix} x(t) + n(t) \end{cases}$$

$$\begin{cases} \dot{z}(t) = \begin{bmatrix} -0.1 & 1 \\ -1 & -0.2 \end{bmatrix} z(t) + m_z(t), \ z(0) = \begin{bmatrix} 0.4 & 0.5 \end{bmatrix}^{\mathrm{T}} \\ \bar{y}(t) = \begin{bmatrix} 0.2 & 4 \\ 2 & 5 \end{bmatrix} z(t) + n_z(t) \end{cases}$$

$$\dot{w}(t) = \begin{bmatrix} -5 & 1 \\ -1 & -9 \end{bmatrix} w(t) + m_w(t), \ w(0) = \begin{bmatrix} 0.3 & 0.4 \end{bmatrix}^{\mathrm{T}}$$

The covariance $Q_m$ and $Q_n$ of $m(t)$ and $n(t)$ are $Q_m = Q_n = 2$, respectively. The covariance $Q_m$ and $Q_n$ of $m_z(t)$ and $n_z(t)$ are $Q_{m_z} = Q_{n_z} = 2$, respectively. And the covariance $Q_w$ of $m_w(t)$ is $Q_w = 2$. The total simulation time is $T = 300(s)$ .we compare the Kalman filter based feedforward-feedback tracking control law to the classical feedforward-feedback tracking control law. The result comparison is showed in Fig.2 to Fig.7.

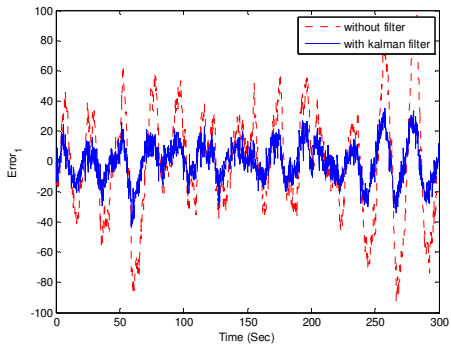

Figure 2. Comparison on error-1

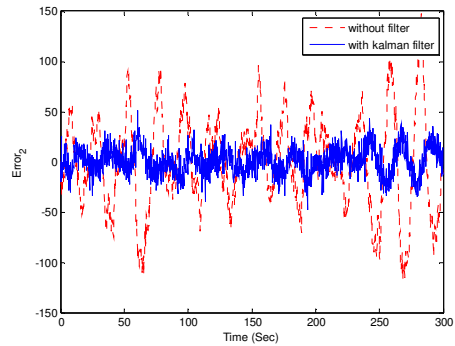

Figure 3. Comparison on error-2

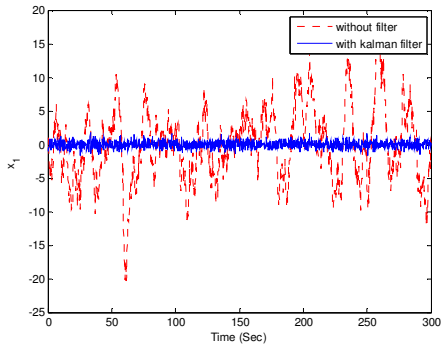

Figure 4. Comparison on state-1

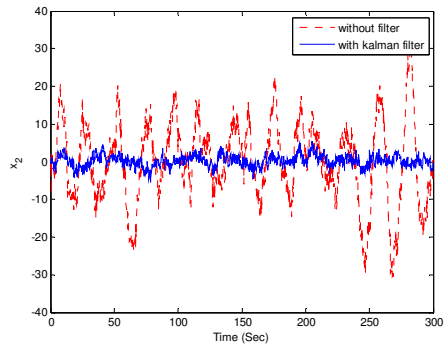

Figure 5. Comparison on stater-2

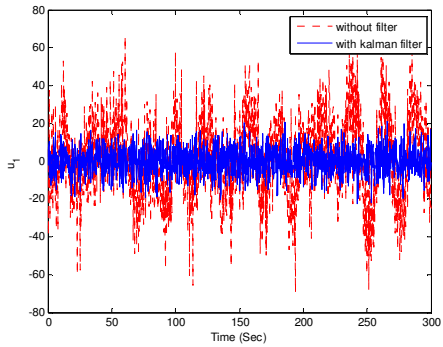
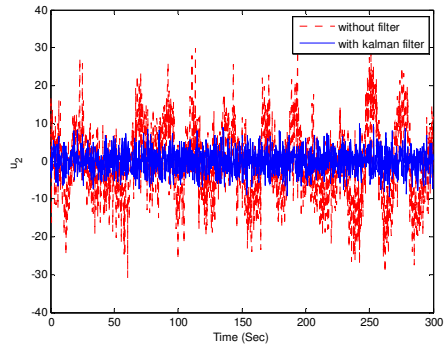

Figure 6. Comparison on control-1       Figure 7. Comparison on control-2

From the simulation result comparison, it is ensured that the Kalman filter based optimal consensus tracking control algorithm is effectiveness. The system tracks the expected the external system(3) in higher precision under the noise and disturbance. The control law is of better noise rejection and the tracking error is smaller than the classic feedforward-feedback tracking control law.

## 5. CONCLUSIONS

Because the system states of consensus protocol is polluted by the noise, we use Kalman filter to filter the noised controlled system to obtain the optimal estimated values of each AUV states, in order to achieve coordination control of multi-AUVs in noisy environment. Considering the controlled system affected by environmental noise and external disturbances, we design a Kalman filter-based feedforward and feedback optimal consensus tracking protocol.

## Authors

Dr. Yuan is Associate Professor, Institute of Oceanographic Instrumentation of Shandong Academy of Science, Qingdao, China. His current research areas are: Multiple Agents System based control, Networked Control System and Nonlinear Filtering. He has published over 20 papers in international journals and international conferences.

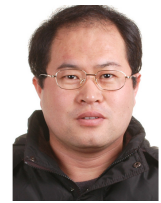